\documentclass[twoside,11pt]{article}

\usepackage{jmlr2e}
\usepackage{amsmath}    
\usepackage{amsfonts}   
\usepackage{amssymb}    
\usepackage{array}    
\usepackage{tabularx}   
\usepackage{times}              
\usepackage{epsfig}
\usepackage{multirow}
\usepackage{subfigure}
\usepackage{algorithm}
\usepackage{algorithmic}
\usepackage{bm}
\usepackage{hyperref}
\usepackage{url}
\usepackage{xspace}

\newcommand{\vv}{{\bf v}}
\newcommand{\hh}{{\bf h}}
\newcommand{\nn}{{\bf n}}

\newcommand{\WW}{{\bf W}}
\newcommand{\VV}{{\bf V}}

\newcommand{\gggg}{{\bf g}}
\newcommand{\bb}{{\bf b}}
\newcommand{\cc}{{\bf c}}
\newcommand{\ppi}{{\bm \pi}}
\newcommand{\trans}{^\top}

\newcommand{\sigm}{\mathrm{sigm}}

\newcommand{\E}{{\sf{E\mspace{-8.7mu} E}}}  
\newcommand{\reals}{\mathbb{R}}

\newcommand{\nade}{NADE\xspace}
\newcommand{\docnade}{DocNADE\xspace}
\newcommand{\ngram}{$n$-gram\xspace}

\jmlrheading{1}{2016}{1-19}{4/00}{10/00}{Stanislas Lauly, Yin Zheng, Alexandre Allauzen and Hugo Larochelle}

\ShortHeadings{\docnade}{Lauly, Zheng, Allauzen \& Larochelle}
\firstpageno{1}

\begin{document}

\title{Document Neural Autoregressive Distribution Estimation}

\author{\name Stanislas Lauly \email Stanislas.Lauly@USherbrooke.ca \\
       \addr D\'epartement d'informatique\\
       Universit\'e de Sherbrooke\\
       Sherbrooke, Qu\'ebec, Canada
       \AND
       \name Yin Zheng \email 
       yin.zheng@hulu.com \\
       \addr Hulu LLC.\\
       Beijing, China
       \AND
       \name Alexandre Allauzen \email 
       allauzen@limsi.fr \\
       \addr LIMSI-CNRS\\
       Universit\'e Paris Sud\\
       Orsay, France
       \AND       \name Hugo Larochelle \email Hugo.Larochelle@USherbrooke.ca \\
       \addr D\'epartement d'informatique\\
       Universit\'e de Sherbrooke\\
       Sherbrooke, Qu\'ebec, Canada}

\editor{Leslie Pack Kaelbling}

\maketitle

\begin{abstract}

We present an approach based on feed-forward neural networks for learning the distribution of textual documents. This approach is inspired by the Neural Autoregressive Distribution Estimator (NADE) model, which has been shown to be a good estimator of the distribution of discrete-valued  high-dimensional vectors. In this paper, we present how NADE can successfully be adapted to the case of textual data, retaining from NADE the property that sampling or computing the probability of observations can be done exactly and efficiently. The approach can also be used to learn deep representations of documents that are competitive to those learned by the alternative topic modeling approaches. Finally, we describe how the approach can be combined with a regular neural network N-gram model and substantially improve its performance, by making its learned representation sensitive to the larger, document-specific context. 

\end{abstract}


\section{Introduction}

One of the most common problem addressed by machine learning is estimating the distribution $p({\bf v})$ of multidimensional data from a set of examples $\{{\bf v}^{(t)}\}_{t=1}^T$. Indeed, good estimates for $p({\bf v})$ implicitly requires modeling the dependencies between the variables in ${\bf v}$, which is required to extract meaningful representations of this data or make predictions about this data.

The biggest challenge one faces in distribution estimation is the well-known curse of dimensionality. In fact, this issue is particularly important in distribution estimation, even more so than in other machine learning problems. This is because a good distribution estimator effectively requires providing an accurate value for $p({\bf v})$ for any value of $\bf v$ (i.e.\ not only for likely values of $\bf v$), with the number of possible values taken by $\bf v$ growing exponentially as the number of the dimensions of the input vector $\bf v$ increases.

One example of a model that has been successful at tackling the curse of dimensionality is the restricted Boltzmann machine (RBM) \citep{Hinton2002}. The RBM and other models derived from it (e.g.\ the Replicated Softmax of \cite{Salakhutdinov-NIPS2010-softmax}) are frequently trained as models of the probability distribution of high-dimensional observations and then used as feature extractors. Unfortunately, one problem with these models is that for moderately large models, calculating their estimate of $p({\bf v})$ is intractable. Indeed, this calculation requires computing the so-called partition function, which normalizes the model distribution. The consequences of this property of the RBM are that approximations must be taken to train it by maximum likelihood and its estimation of $p({\bf v})$ cannot be entirely trusted.

In an attempt to tackle these issues of the RBM, the Neural Autoregressive Distribution Estimator (NADE) was introduced by \cite{Larochelle+Murray-2011}. \nade's parametrization is inspired by the RBM, but uses feed-forward neural networks and the framework of autoregression for modeling the probability distribution of binary variables in high-dimensional vectors. Importantly, computing the probability of an observation under \nade can be done exactly and efficiently.

In this paper, we describe a variety of ways to extend \nade to model data from text documents. 
We start by describing Document \nade (\docnade), a single hidden layer feed-forward neural network model for bag-of-words observations, i.e.\ orderless sets of words. This requires adapting NADE to vector observations $\bf v$, where each of element $v_i$ represents a word and where the order of the dimensions is random. Each word is represented with a lower-dimensional, real-valued embedding vector, where similar words should have similar embeddings. This is in line with much of the recent work on using feed-forward neural network models to learn word vector embeddings~\citep{bengio2003neural,mnih2007three,Mnih+Hinton-2009,MikolovT2013} to counteract the curse of dimensionality. However, in \docnade, the word representations are trained to reflect the topics (i.e.\ semantics) of documents only, as opposed to their syntactical properties, due to the orderless nature of bags-of-words.

Then, we describe how to train deep versions of \docnade. First described by \cite{zheng2014deep} in the context of image modeling, here we empirically evaluate them for text documents and show that they are competitive to alternative topic models, both in terms of perplexity and document retrieval performances.

Finally, we present how the topic-level modeling ability of \docnade can be used to obtain a useful representation of context for language modeling. We empirically demonstrate that by learning a topical representation of previous sentences, we can improve the perplexity performance of an N-gram neural language model.



\section{Document NADE (\docnade)}
    \docnade is derived from the Neural Autoregressive Distribution Estimation (\nade) that will be first described in this section. Implemented as a feed-forward architecture, it extends NADE to provide an efficient and meaningful generative model of document bags-of-words. 

    \subsection{Neural Autoregressive Distribution Estimation (\nade)}
    \nade, introduced in~\cite{Larochelle+Murray-2011}, is a tractable distribution estimator for modeling the distribution of high-dimensional vectors of binary variables. Let us consider a binary vector of $D$ observations, $\vv \in \{0,1\}^D$. The \nade model estimates the probability of this vector by applying the probability chain rule as follows:
    \begin{align}\label{eq:nade-chain}
    p(\vv) = \prod_{i=1}^D p(v_i|\vv_{<i}),
    \end{align}
    where $v_i$ denotes the $i$-th component of $\vv$ and $\vv_{<i} \in  \{0,1\}^{i-1}$ contains the first $i-1$ components of $\vv$: $\vv_{<i}$ is the sub-vector $[v_1,\dots,v_{i-1}]\trans$. The peculiarity of \nade lies in the neural architecture designed to estimate the conditional probabilities involved in Equation~\ref{eq:nade-chain}. 
    To predict the component $i$, the model first computes its hidden layer of dimension $H$
\begin{align}\label{eq:nade-hidden}
\hh_i(\vv_{<i})  = \gggg\left(\cc + \WW_{:,<i} \vv_{<i} \right),
\end{align}
leading to the following probability model:
\begin{align}
p(v_i=1|\vv_{<i}) = \sigm\left(b_i + \VV_{i,:} \hh_i(\vv_{<i}) \right). \label{eqn:nade-prob}
\end{align}
In these two equations,  $\sigm(x) = 1/(1+\exp(-x))$ denotes the sigmoid activation function while
function $\gggg(\cdot)$ could be any activation function, though \cite{Larochelle+Murray-2011} also used the sigmoid function. $\WW \in \reals^{H\times D}$ and $\VV \in \reals^{D\times H}$ are the parameter matrices along with the associated bias terms $\bb \in \reals^{D}$ and $\cc \in \reals^{H}$, with
$\WW_{:,<i}$ being a matrix made of the $i-1$ first columns of $\WW$.

\begin{figure}[t]
\vspace*{-.5cm}
\begin{center}
\includegraphics[width=4.0cm]{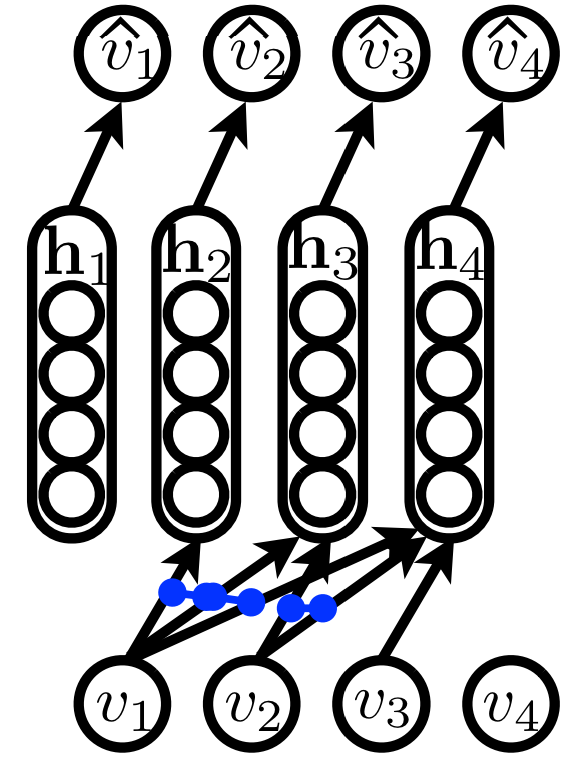}
\end{center}
\vspace*{-.5cm}
\caption{Illustration of \nade. Colored lines
  identify the connections that share parameters and
  $\widehat{v}_i$ is a shorthand for the autoregressive
  conditional $p(v_i|\vv_{<i})$.  The observations $v_i$ are binary.}
\label{fig:models}
\end{figure}

Instead of a single projection of the input vector, the \nade model relies on a set of separate  hidden layers $\hh_i(\vv_{<i})$ that each represent the previous inputs in a latent space. The connections between input dimension $v_i$ and each hidden layer $\hh_i(\vv_{<i})$ are tied as shown in figure~\ref{fig:models}, allowing the model to compute all the hidden layers for one input in $O(D H)$. The parameters $\{\bb,\cc,\WW,\VV\}$ are learned by minimizing the average negative log-likelihood using stochastic gradient descent.

\subsection{From \nade to \docnade}
        
    The Document \nade model (\docnade) aims at learning meaningful representations of texts from an unlabeled collection of documents. This model embeds, like \nade, a set of hidden layers. Their role is to capture salient statistical patterns in the co-occurrence of words within documents and can be considered as modeling hidden topics. 
    
    To represent a document, the vector $\vv$ is now a sequence of arbitrary size $D$. Each element of $\vv$ corresponds to a multinomial observation over a fixed vocabulary of size $V$. Therefore $v_i \in \{1, ..., V\}$ represents the index in the vocabulary of the $i$-th word of the document. For now, we'll assume that an ordering for the words is given, but we will discuss the general case of orderless bags-of-words in Section~\ref{ssec:training-from-BOW}.
        
    The main approach taken by \docnade is similar to \nade, but differs significantly in the design of parameter tying. The probability of a document $\vv$ is estimated using the probability chain rule, but the architecture is modified to cope with large vocabularies.
    Each word observation $v_i$ of the document $\vv$ leads to a hidden layer $\hh_i$, which represents the past observations $\vv_{<i}$. This hidden layer is computed as follows: 
    \begin{align}\label{eq:docnade-h}
    \hh_{i}(\vv_{<i}) = \gggg\left( \cc + \sum_{k<i} \WW_{:,v_k} \right),
    \end{align}
    where each column of the matrix $\WW$ acts as a vector of size $H$ that represents a word. The embedding of the $i^{th}$ word in the document is thus the column of index $v_i$ in the matrix $\WW$. 
    
    Notice that by sharing the word representation matrix across positions in the document, each hidden layer $hh_{i}(\vv_{<i})$ is in fact independent of the order of the words within $\vv_{<i}$. The implications of this choice is that the learned hidden representation will not model the syntactic structure of the documents and focus on its document-level semantics, i.e.\ its topics.
    
    It is also worth noticing that we can compute $\hh_{i+1}$ recursively by keeping track of the pre-activation of the previous hidden layer $\hh_i$ as follows: 
\begin{align}
  \hh_{i+1}(\vv_{<i+1}) = \gggg\left(\WW_{:,v_i} + \underbrace{\cc + \sum_{k<i} \WW_{:,v_k}}_{\text{Precomputed for} ~\hh_{i}(\vv_{<i})} \right) \label{eqn:efficientHiddens}
\end{align}
    The weight sharing between hidden layers enables us to compute all hidden layers $\hh_i(\vv_{<i})$ for a document in $O(D H)$. 
    
    Then, to compute the probability of a full document $p(\vv)$, we need to estimate all  conditional probabilities $p(v_i|\vv_{<i})$. A straightforward solution would be to compute each $p(v_i|\vv_{<i})$ using softmax layers with a shared weight matrix and bias, each fed with the  corresponding hidden layer $\hh_{i}$. However, the computational cost of this approach is prohibitive, since it scales linearly with the vocabulary size\footnote{For most natural language processing task, the vocabulary size exceeds $10,000$.}.To overcome this issue, we represent distribution over the vocabulary by a probabilistic binary tree, where the leaves correspond to the words. This approach is widely used in the field of neural probabilistic language models~\citep{Morin+al-2005,Mnih+Hinton-2009}. Each word is represented by a path in the tree, going from the root to the leaf associated to that word. A binary logistic regression unit is associated to each node in the tree and gives the probability of the binary choice, going left or right.
    Therefore, a word probability can be estimated by the path's probability in this tree, resulting in a complexity in $O(\log V)$ for trees that are balanced. In our experiments, we used a randomly generated full binary tree with $V$ leaves, each
    assigned to a unique word of the vocabulary. An even better option would be to derive the tree using Hoffman coding, which would reduce even more the average path lengths.
    
    More formally, let's denote by ${\bf l}(v_i)$ the sequence of nodes composing the path, from the root of the tree to the leaf corresponding to word $v_i$. Then, $\ppi(v_i)$ is the sequence of left/right decisions of the nodes in ${\bf l}(v_i)$. For example, the root of the tree is always the first element $l(v_i)_1$ and the value $\pi(v_i)_1$ will be 0 if the word is in the left sub-tree  and 1 if it is in the right sub-tree. The matrix $\bf V$ stores by row the weights associated to each logistic classifier. There is one logistic classifier per node in the tree. Let $\VV_{l(v_i)_m,:}$ and $b_{l(v_i)_m}$ be the weights and bias for the logistic unit associated to the node $n(v_i)_m$. The probability $p(v_i|\vv_{<i})$ given the tree and the hidden layer $\hh_i(\vv_{<i})$ is computed with the following formulas:
\begin{align}
  p(v_i=w|\vv_{<i}) = \prod_{m=1}^{|\ppi(v_i)|} p(\pi(v_i)_m|\vv_{<i})\label{eqn:mnade-tree1}, \textrm{ with}
\end{align}
\begin{align}
  p(\pi(v_i)_m=1|\vv_{<i}) = \sigm\left(b_{l(v_i)_m} + \VV_{l(v_i)_m,:} \hh_i(\vv_{<i})\right)\label{eqn:mnade-tree2}.
\end{align}
    This hierarchical architecture allows us to efficiently compute the probability for each word in a document and therefore the probability of every documents with the probability chain rules (see Equation~\ref{eq:nade-chain}). 
    As in the \nade model, the parameters of the model  $\{\bb,\cc,\WW,\VV\}$ are learnt by minimizing the negative log-likelihood using stochastic gradient descent. 
    
    \begin{figure}[t]
    \vspace*{-.5cm}
    \begin{center}
    \includegraphics[width=8.0cm]{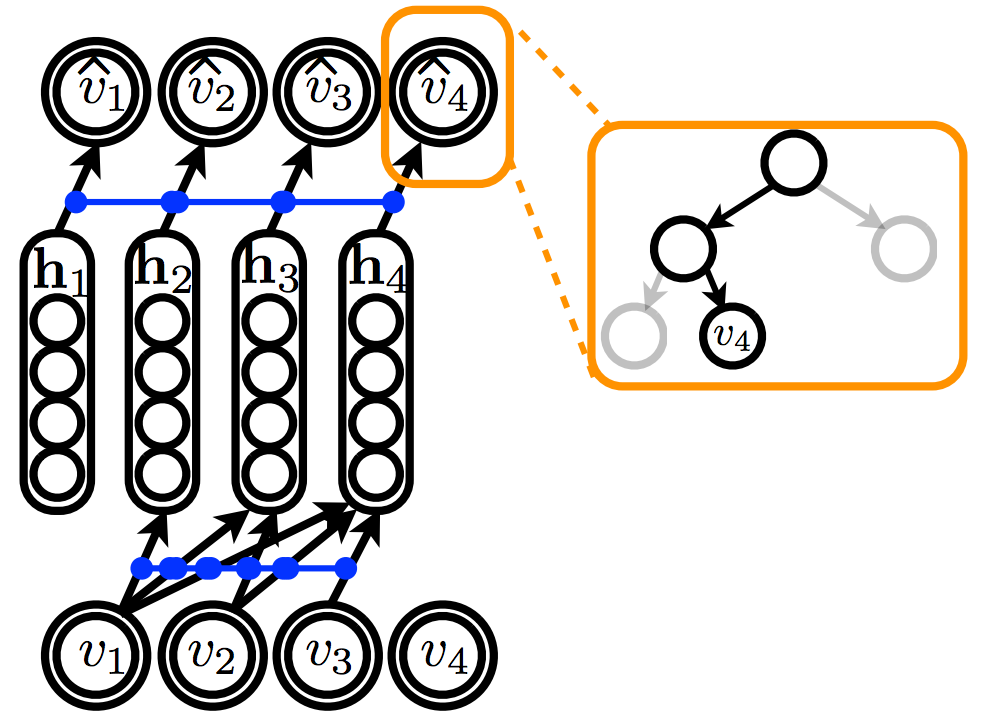}
    \end{center}
    \vspace*{-.5cm}
    \caption{ Illustration of \docnade. Connections between each multinomial observation $v_i$ and hidden units are also shared, and each conditional $p(v_i|\vv_{<i})$ is decomposed into a tree of binary logistic regressions.}
    \label{fig:DocNADE_model}
    \end{figure}
    
    Since there is $\log(V)$ logistic regression units for a word (one per node), each of them has a time complexity of $O(H)$, the complexity of computing the probability for a document of $D$ words is in  $O(\log(V) D H)$. 
    
    As for using \docnade to extract features from a complete document, we propose to use
    \begin{align}
    \hh(\vv) = \hh_{D+1}(\vv_{<D+1}) = \gggg\left( \cc + \sum_{k=1}^D \WW_{:,v_k} \right)
    \end{align}
    which would be the hidden layer computed to obtain the conditional probability of a $D+1^{\rm th}$ word appearing in the document. 
    
    \subsection{Training from bag-of-word counts}
    \label{ssec:training-from-BOW}
    So far, we have assumed that the ordering of the words in the document is known. However, document datasets often take the form of word-count vectors in which the original word order, required to specify the sequence of conditionals $p(v_i|\vv_{<i})$, has been lost. 
    
    Thankfully, it is still possible to successfully train \docnade despite the absence of this information. The idea is to assume that each observed document $\vv$ was generated by initially sampling a {\it seed} document $\widetilde{\vv}$ from \docnade, whose words were then shuffled using a randomly generated ordering to produce $\vv$. With this approach, we can express the probability distribution of $\vv$ by computing the marginal over all possible seed document:
    \begin{align}
        p(\vv) = \sum_{\widetilde{\vv}\in {\cal V}(\vv)} p(\vv,\widetilde{\vv})
        = \sum_{\widetilde{\vv}\in {\cal V}(\vv)} p(\vv|\widetilde{\vv})p(\widetilde{\vv}) \label{eqn:marg}
    \end{align}
    where $p(\widetilde{\vv})$ is modeled by \docnade, $\widetilde{\vv}$ is the same as the observed document $\vv$ but with a different (random) word sequence order and ${\cal V}(\vv)$ is the set of all the documents $\widetilde{\vv}$ that have the same word count $\nn(\vv) = \nn(\widetilde{\vv})$. With the assumption of orderings being uniformly sampled, we can replace $p(\vv|\widetilde{\vv})$ with $\frac{1}{|{\cal V}(\vv)|}$ giving us:
    \begin{align}
        p(\vv) = \sum_{\widetilde{\vv} \in {\cal V}(\vv)} \frac{1}{|{\cal V}(\vv)|} p(\widetilde{\vv})
        = \frac{1}{|{\cal V}(\vv)|}\sum_{\widetilde{\vv} \in {\cal V}(\vv)} p(\widetilde{\vv})~.\label{eqn:pvcounts}
    \end{align}

    In practice, one approach to training the \docnade model over $\widetilde{\vv}$ is to artificially generate ordered documents by uniformly sampling words, without replacement, from the bags-of-words in the dataset. This would be equivalent to taking each original document and shuffling the order of its words. This approach can be shown to optimize a stochastic upper bound on the actual negative log-likelihood of documents. As we'll see, experimental results show that convincing performance can still be reached. 
    
    With this training procedure, \docnade shows its ability to learn and predict a new word in a document at a random position while preserving the overall semantic properties of the document. The model is therefore learning not to insert “intruder” words, i.e.\ words that do not belong with the others. After training, a document's learned representation should contain valuable information to identify intruder words for this document. It's interesting to note that the detection of such intruder words has been used previously as a task in user studies to evaluate the quality of the topics learned by LDA, though at the level of single topics and not whole documents~\citep{ChangJ2009}.

\section{Deep Document NADE}
The single hidden layer version of \docnade already achieves very competitive performance for topic modeling~\citep{larochelle2012neural}. Extending it to a deep, multiple hidden layer architecture could however allow for even better performance, as suggested by the recent and impressive success of deep neural networks. Unfortunately, deriving a deep version of \docnade that is practical cannot be achieved solely by adding hidden layers to the definition of the conditionals $p(v_i=1|\vv_{<i})$. Indeed, computing $p(\vv)$ requires computing each $p(v_i=1|\vv_{<i})$ conditional (one for each word), and it is no longer possible to exploit Equation~\ref{eqn:efficientHiddens} to compute the sequence of all hidden layers in $O(D H)$ when multiple deep hidden layers are used. 

In this section, we describe an alternative training procedure that enables us the introduction of multiple stacked hidden layers. This procedure was first introduced in~\cite{zheng2014deep} to model images, which was itself borrowing from the training scheme 
introduced by \cite{Uria2013b}. 

As mentioned in section~\ref{ssec:training-from-BOW}, \docnade can be trained on random permutations of the words from training documents. As noticed by~\cite{Uria2013b},
the use of many orderings during training can be seen as the instantiation of many different \docnade models that share a single set of parameters. Thus, training \docnade with random permutations also amounts to minimizing the negative log-likelihood averaged across all possible orderings, for each training example $\vv$.

In the context of deep NADE models, a key observation is that training on all possible orderings implies that for a given context $\vv_{<i}$, we wish the model to be equally good at predicting any of the remaining words appearing next, since for each there is an ordering such that they appear at position $i$. 
Thus, we can redesign the training algorithm such that, instead of sampling a complete ordering of all words for each update, we instead sample a single context $\vv_{<i}$ and perform an update of the conditionals using that context. This is done as follows. For a given document, after generating vector $\vv$ by shuffling the words from the document, a split point $i$ is randomly drawn. From this split point results two parts of the document: $\vv_{<i}$ and  $\vv_{\geq i}$. The former is considered as the input and the latter contains the targets to be predicted by the model. Since in this setting a training update relies on the computation of a single latent representation, that of $\vv_{<i}$ for the drawn value of $i$, deeper hidden layers can be added at a reasonable increase in computation.

Thus, in Deep\docnade the conditionals $p(v_{i} | \vv_{<i})$ are modeled as follows. The first hidden layer $\hh^{(1)}(\vv_{<i})$ represents the conditioning context $\vv_{<i}$ as in the single hidden layer \docnade:

\begin{eqnarray}
\mathbf{h}^{\left(1\right)}\left({\bf v}_{<i}\right) &= {\bf g}\left( \mathbf{c}^{\left(1\right)}+\sum_{k<i}\mathbf{W}^{\left(1\right)}_{:,v_{k}} \right ) = {\bf g}\left({\bf c}^{\left(1\right)} + {\bf W}^{\left(1\right)}\mathbf{x}\left({\bf v}_{<i}\right) \right) \label{eqn: deepdocnade_h1}
\end{eqnarray}
where $\mathbf{x}\left({\bf v}_{<i}\right)$ is the histogram vector representation
of the word sequence ${\bf v}_{<i}$, and
the exponent is used as an index over the hidden layers and its parameters, with $(1)$ referring to the first layer. We can now easily add new hidden layers as in a regular deep feed-forward neural network:
\begin{align}
\hh^{(n)}(\vv_{<i}) = \gggg (\cc^{(n)} +  W^{(n)} \hh^{(n-1)}(\vv_{<i})),
\end{align}
for $n = 2, \dots , N$, where $N$ is the total number of hidden layers. From the last hidden layer $\hh^{N}$, we can finally compute the conditional $p(v_i=w | \vv_{<i})$, for any word $w$.

Finally, the loss function used to update the model for the given context $\vv_{<i}$ is:
\begin{align}
L(\vv) = \frac{D_{\vv}}{D_{\vv} - i + 1} \sum_{w \in \vv_{\geq i}} - \log p(v_i = w | \vv_{<i}),\label{eqn:loss}
\end{align}
where $D_{\vv}$ is the number of words in $\vv$ and the sum iterates over all words $w$ present in $\in \vv_{\geq i}$. Thus, as described earlier, the model predicts each remaining word after the splitting position $i$ as if it was actually at position $i$. The factors in front of the sum comes from the fact that the complete log-likelihood would contain $D_{\vv}$ log-conditionals and that we are averaging over $D_{\vv} - i + 1$ possible choices for the word ordered at position $i$. For a more detailed presentation, see \cite{zheng2014deep}. The average loss function of Equation~\ref{eqn:loss} is optimized with stochastic gradient descent\footnote{A document is usually represented as bag-of-words. Generating a word vector $\vv$ from its bag-of-words, shuffling the word count vector $\vv$, splitting it, and then regenerating the histogram $\mathbf{x}\left({\bf v}_{<i}\right)$ and $\mathbf{x}\left({\bf v}_{\ge i}\right)$ is unfortunately fairly inefficient for processing samples in a mini-batch fashion. Hence, in practice, we split the original histogram $\mathbf{x}\left({\bf v}\right)$ directly by uniformly sampling, for each word individually, how many are put on the left of the split (the others are put on the right of the split). This procedure, used also by \cite{zheng2014deep}, is only an approximation of the correct procedure mentioned in the main text, but produces a substantial speedup while also yielding good performance. Thus, we used it also in this paper.}.

Note that to compute the probabilities $p(v_i = w | \vv_{<i})$, a probabilistic tree could again be used. However, since all probabilities needed for an update are based on a single context $\vv_{<i}$, a single softmax layer is sufficient to compute all necessary quantities. Therefore the computational burden of a conventional softmax is not as prohibitive as for \docnade, especially with an efficient implementation on the GPU. For this reason, in our experiments with Deep\docnade we opted for a regular softmax.

\section{\docnade Language Model}
\label{sec:docnade-lm}
While topic models such as \docnade can be useful to learn topical representations of documents, they are actually very poor models of language. In \docnade, this is due to the fact that, when assigning a probability to the next word in a sentence, \docnade actually ignores in which order the previously observed words appeared. Yet, this ordering of words conveys a lot of information regarding the most likely syntactic role of the next word or the finer semantics within the sentence. In fact, most of that information is predictable from the last few words, which is why N-gram language models remain the dominating approach to language modeling. 

In this section, we propose a new model that extends \docnade to mitigate the influence of both short and long term dependencies in a single model, which we refer to as the \docnade language model or \docnade-LM. The solution we propose enhances the bag-of-word representation of a word's history with the explicit inclusion of \ngram dependencies for each word to predict. 

\begin{figure}[t]
    \vspace*{-.5cm}
    \begin{center}
    \includegraphics[width=10.0cm]{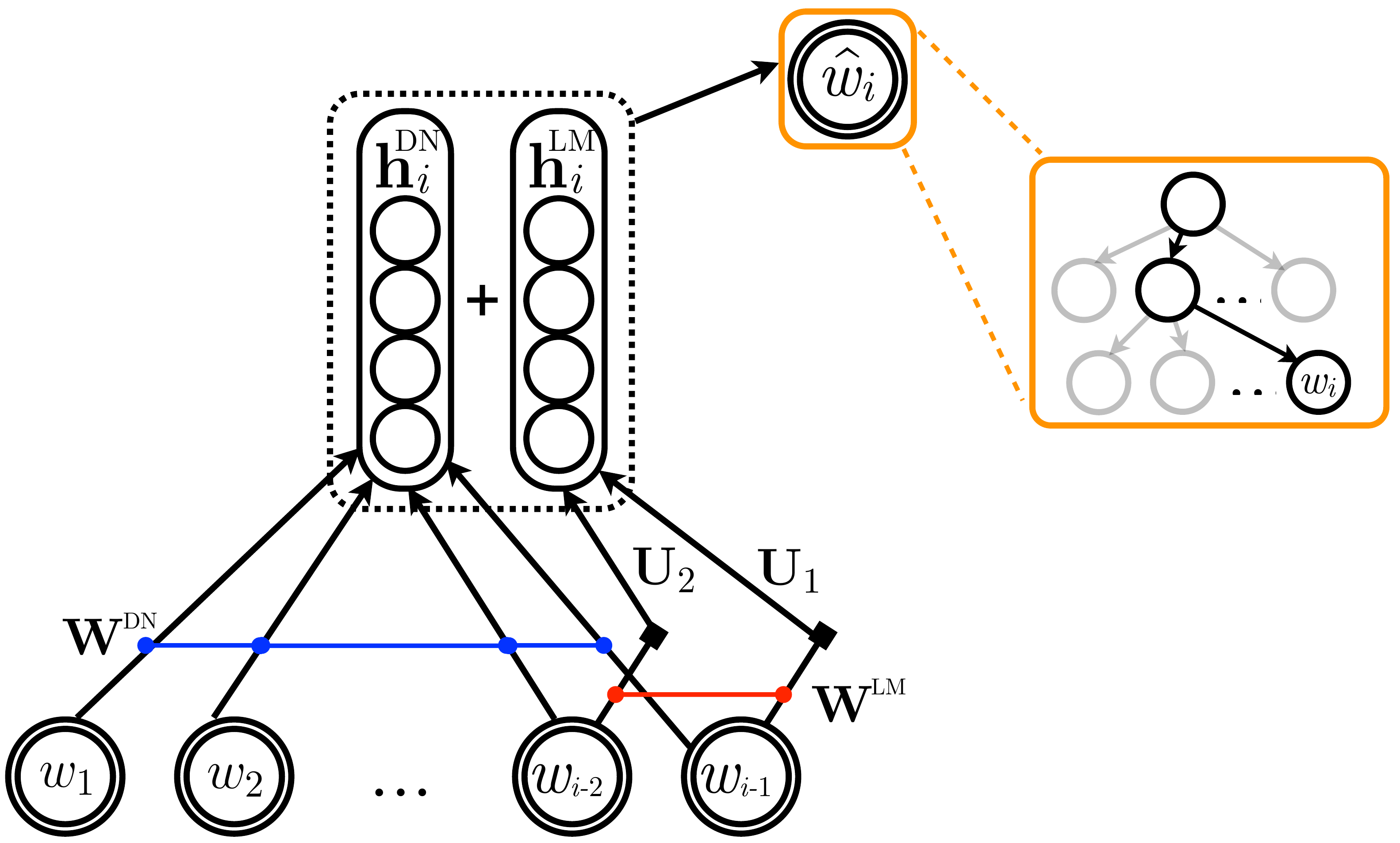}
    \end{center}
    \vspace*{-.5cm}
    \caption{ Illustration of the conditional $p(v_i|\vv_{<i})$ in a trigram NADE language model. Compared to \docnade, this model incorporates the architecture of a neural language model, that first maps previous words (2 for a trigram model) to vectors using an embedding matrix ${\bf W}^{\rm LM}$ before connecting them to the hidden layer using regular (untied) parameter matrices (${\bf U_1}$, ${\bf U_2}$ for a trigram). In our experiments, each conditional $p(v_i|\vv_{<i})$ exploits decomposed into a tree of logistic regressions, the hierarchical softmax.}
    \label{fig:NADE_lm}
    \end{figure}

The figure~\ref{fig:NADE_lm} depicts the overall architecture. This model can be seen as an extension of the seminal work on neural language models of \cite{bengio2003neural} that includes a representation of a document's larger context. It can also be seen as a neural extension of the cache-based language model introduced in~\citep{kuhn90}, where the \ngram probability is interpolated with the word distribution observed in a dynamic cache. This cache of a fixed size keeps track of previously observed words to include long term dependencies in the prediction and preserve semantic consistency beyond the scope of the \ngram. In our case, the \docnade language model maintains an unbounded cache and define a proper, jointly trained solution to mitigate these two kinds of dependencies. 


As in \docnade, a document $\vv$ is modeled as a sequence of multinomial observations. The sequence size is arbitrary and each element $v_i$ consists in the index of the $i$-th word in a vocabulary of size $V$. 
The conditional probability of a word given its history $p(v_i|\vv_{<i})$ is now expressed as a smooth function of a hidden layer $\hh_i(\vv_{<i})$ used to predict word $v_i$. The peculiarity of the \docnade language model lies in the definition of this hidden layer, which now includes two terms: 
\begin{align}\label{eq:docnadelm-hidden}
    \hh_i(\vv_{<i}) = \gggg(\bb + {\bf h}^{\rm DN}_i(\vv_{<i}) + {\bf h}^{\rm LM}_i(\vv_{<i})).
\end{align}
The first term borrows from the \docnade model by aggregating embeddings for all the previous words in the history: 
\begin{align}\label{eq:docnadelm-dn}
    {\bf h}^{\rm DN}_i(\vv_{<i}) = \sum_{k<i} \WW^{\rm DN}_{:,v_k}~.
\end{align}
The second contribution derives from neural \ngram language models as follows: 
\begin{align}
    {\bf h}^{\rm LM}_i(\vv_{<i}) = \sum_{k=1}^{n-1} {\bf U}_k \cdot \WW^{\rm LM}_{:,v_{i-k}}~. \label{eqn:FFN}
\end{align}
In this formula, the history for the word $v_i$ is restricted to the $n-1$ preceding words, following the common \ngram assumption. The term ${\bf h}^{\rm LM}_i$ hence represents the continuous representation of these $n-1$ words, in which word embeddings are linearly transformed by the ordered matrices ${\bf U}_1, {\bf U}_2, ..., {\bf U}_{n-1}$. 
Moreover, $\bb$ gathers the bias terms for the hidden layer. In this model, two sets of word embeddings are defined,  $\WW^{\rm DN}$ and  $\WW^{\rm LM}$, which are respectively associated to the \docnade and neural language model parts. For simplicity, we assume both are of the same size $H$.

Given hidden layer $\hh_i(\vv_{<i})$, conditional probabilities $p(v_ i | \vv_{<i})$ can be estimated, and thus $p(\vv)$. For the aforementioned reason, the output layer is structured for efficient computations. 
Specifically, we decided to use a variation of the probabilistic binary tree, known as a hierarchical softmax layer. In this case, instead of having binary nodes with multiple levels in the tree, we have only two levels where all words have their leaf at level two and each node is a multiclass (i.e.\ softmax) logistic regression with roughly $\sqrt{V}$ classes (one for each children). Computing probabilities in such a structured layer can be done using only two matrix multiplications, which can be efficiently computed on the GPU.


%




With a hidden layer of size $H$, the complexity of computing the softmax at one node is $O(H \sqrt{V})$. If we have $D$ words in a given document, the complexity of computing all necessary probabilities from the hidden layers is thus $O(D H \sqrt{V})$. It also requires $O(D H)$ computations to compute the hidden representations for the \docnade part and $O(n H^2)$ for the language model part. The full complexity for computing $p(\vv)$ and the updates for the parameters are thus computed in $O(D H \sqrt{V} + D H + n H^2)$.

Once again, the loss function of the model is the negative log-likelihood and we minimize it by using stochastic gradient descent over documents, to learn the values of the parameters $\allowbreak \{\bb, \cc, \VV, \WW^{\rm LM}, \WW^{\rm DN}, {\bf U}_1, ..., {\bf U}_n\}$.

\section{Related Work}

Much like NADE was inspired by the RBM, \docnade can be seen as related to the Replicated Softmax model \citep{Salakhutdinov-NIPS2010-softmax}, an extension of the RBM to document modeling. Here, we describe in more detail the Replicated Softmax, along with its relationship with \docnade.

\begin{figure}[t]
    \vspace*{-.5cm}
    \begin{center}
    \includegraphics[width=4.0cm]{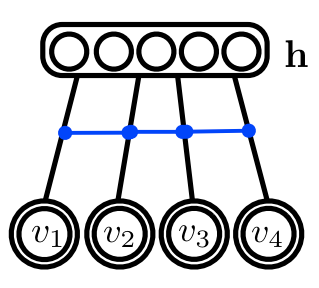}
    \end{center}
    \vspace*{-.5cm}
    \caption{Replicated Softmax model. Each multinomial observation $v_i$ is a word. Connections between each multinomial observation $v_i$ and hidden units are shared.}
    \label{fig:repSoft}
    \end{figure}
    

Much like the RBM, the Replicated Softmax models observations using a latent, stochastic binary layer $\hh$. Here, the observations are the documents $\vv$, which interact with the hidden layer $\hh$ through an energy function similar to RBM's:
\begin{align}
  E(\vv,\hh) &= - D~\cc\trans \hh + \sum_{i=1}^D  -\hh\trans \WW_{:,v_i}  -b_{v_i}
  = - D~\cc\trans \hh -\hh\trans \WW \nn(\vv) -\bb\trans \nn(\vv),
\end{align}
where $\nn(\vv)$ is a bag-of-word vector of size $V$ (the size of the vocabulary) containing the word count of each word in the vocabulary for document $\vv$. $\bf h$ is the stochastic, binary hidden layer vector and  $\WW_{:,v_i}$ is the $v_i^{\rm th}$ column vector of matrix $\WW$. $\bf c$ and $\bf b$ are the bias vectors for the visible and the hidden layers. We see here that the larger $\bf v$ is, the bigger the number of terms in the sum over $i$ is, resulting in a high energy value.
For this reason, the hidden bias term $\cc\trans \hh$ is multiplied by $D$, to be commensurate with the contribution of the visible layer..
We can see also that connection parameters are shared across different positions $i$ in $\vv$, as illustrated by figure \ref{fig:repSoft}). 
    
The conditional probabilities of the hidden and the visible layer factorize much like in the RBM, in the following way:
\begin{align}
p(\hh|\vv) = \prod_j p(h_j| \vv)~,~~~~~
p(\vv|\hh) = \prod_{i=1}^D p(v_i | \hh)
\end{align}
where the factors $p(h_j| \vv)$ and $p(v_i | \hh)$ are such that
\begin{align}
   p(h_j=1|\vv) &= \sigm(D c_j + \sum_i W_{jv_i})\label{eqn:rp-gibbs-v}\\
   p(v_i = w |\hh) &= \frac{\exp(b_w + \hh\trans \WW_{:,w})}{\sum_{w'}\exp(b_{w'} + \hh\trans \WW_{:,w'})}\label{eqn:rp-gibbs-h}
\end{align}
The normalized exponential part in $p(v_i = w |\hh)$ is simply the softmax function. To train this model, we'd like to minimize the negative log-likelihood (NLL). Its gradients for a document ${\bf v}^{(t)}$ with respect to the parameters $\theta = \{{\bf W}, {\bf c}, {\bf b}\}$ are calculated as follows:
\begin{align}
\frac{\partial -\log p(\vv^{(t)})}{\partial \theta} & = \E_{\hh | \vv^{(t)}}\left[\frac{\partial}{\partial \theta} E(\vv^{(t)},\hh)\right] - \E_{\vv,\hh}\left[\frac{\partial}{\partial \theta} E(\vv,\hh)\right].\label{eqn:grad}
\end{align}
As with conventional RBMs, the second expectation in Equation~\ref{eqn:grad} is computationally too expensive. The gradient of the negative log-likelihood is therefore approximated by replacing the second expectation with an estimated value obtained by contrastive divergence~\citep{Hinton2002}. This approach consists of performing $K$ steps of blocked Gibbs sampling, starting at $\vv^{(t)}$ and using Equations \ref{eqn:rp-gibbs-v} and \ref{eqn:rp-gibbs-h}, to obtain point estimates of the expectation over $\vv$. Large values of $K$ must be used to reduce the bias of gradient estimates and obtain good estimates of the distribution. This approximation is used to perform stochastic gradient descent. 

During Gibbs sampling, the Replicated Softmax model must compute and sample from $p(v_i = w |\hh)$, which requires the computation of a large softmax. Most importantly, the computation of the softmax most be repeated $K$ times for each update, which can be prohibitive, especially for large vocabularies. Unlike \docnade, this softmax cannot simply be replaced by a structured softmax.


It is interesting to see that \docnade is actually related to how the Replicated Softmax approximates, through mean-field inference, the conditionals $p(v_i=w|\vv_{<i})$. Computing the conditional $p(v_i=w|\vv_{<i})$ with the Replicated Softmax is intractable. However, we could use mean-field inference to approximate the full conditional $p(v_i,\vv_{>i},\hh|\vv_{<i})$ as the factorized
\begin{align}
q(v_i,\vv_{>i},\hh|\vv_{<i}) = \prod_{k\geq i} q(v_k|\vv_{<i}) \prod_j q(h_j|\vv_{<i})~,
\end{align}
where $q(v_k=w|\vv_{<i}) = \mu_{kw}(i)$ and $q(h_j=1|\vv_{<i}) = \tau_j(i)$. We would find the parameters $\mu_{kw}(i)$ and $\tau_j(i)$ that minimize the KL divergence between $q(v_i,\vv_{>i},\hh|\vv_{<i})$ and $p(v_i,\vv_{>i},\hh|\vv_{<i})$ by applying the following message passing equations until convergence:
 \begin{align}
&\tau_j(i) \leftarrow \sigm\left(D~c_j + \sum_{k\geq i} \sum_{w'=1}^V W_{jw'} \mu_{kw'}(i) + \sum_{k<i} W_{jv_k}  \right), \\
&\mu_{kw}(i) \leftarrow \frac{\exp(b_w + \sum_{j} W_{jw} \tau_j(i))}{\sum_{w'}\exp(b_{w'} + \sum_{j} W_{jw'} \tau_j(i))}. 
\end{align}
with $k\in\{i,\dots,D\}$, $j \in\{1,\dots,H\}$ and $w\in\{1,\dots,V\}$. The conditional $p(v_i=w|\vv_{<i})$ could then be estimated with $\mu_{kw}(i)$ for all $i$. We note that one iteration of mean-field (with $\mu_{kw'}(i)$ initialized to 0) in the Replicated Softmax corresponds to the conditional $p(v_i=w|\vv_{<i})$ computed by \docnade with a single hidden layer and a flat softmax output layer.


In our experiment, we'll see that \docnade compares favorably to Replicated Softmax.

\section{Topic modeling experiments}

To compare the topic models, two kinds of quantitative comparisons are used. The first one evaluates  the generative ability of the different models, by computing the perplexity of held-out texts. The second one compares the quality of document representations for an information retrieval task.

Two different datasets are used for the experiments of this section, 20~Newsgroups and RCV1-V2 (Reuters Corpus Volume I). The 20~Newsgroups corpus has 18,786 documents (postings) partitioned into 20 different classes (newsgroups). RCV1-V2 is a much bigger dataset composed of 804,414 documents (newswire stories) manually categorized into 103 classes (topics). The two datasets were preprocessed by stemming the text and removing common stop-words. The 2,000 most frequent words of the 20~Newsgroups training set and the 10,000 most frequent words of the RCV1-V2 training set were used to create the dictionary for each dataset. Also, every word counts $n_i$, used to represent the number of times a word appears in a document was replaced by $log(1 + n_i)$ rounded to the nearest integer, following \cite{Salakhutdinov-NIPS2010-softmax}.

\subsection{Generative Model Evaluation}
For the generative model evaluation, we follow the experimental setup proposed by \citet{Salakhutdinov-NIPS2010-softmax} for 20~Newsgroups and RCV1-V2 datasets. We use the exact same split for the sake of comparison. The setup consists in respectively 11,284 and 402,207 training examples for 20~Newsgroups and  RCV1-V2. We randomly extracted 1,000 and 10,000 documents from the training sets of 20 Newsgroups and RCV1-V2, respectively, to build a validation set. 
The average perplexity per word is used for comparison. This perplexity is estimated using the 50 first test documents, as follows: 
\begin{align}
    \exp\left(- \frac{1}{T}\sum_t \frac{1}{|\vv^t|}\log p(\vv^t)\right),
\end{align}
where $T$ is the total number of examples and $\vv^t$ is the $t^{\rm th}$ test document\footnote{
Note that there is a difference between the sizes, for the training sets and test sets of 20~Newsgroups and RCV1-V2 reported in this paper and the one reported in the original data paper of \citet{Salakhutdinov-NIPS2010-softmax}. The correct values are the ones given in this section, which was confirmed after personal communication with \citet{Salakhutdinov-NIPS2010-softmax}.}.


\begin{table}[h]
\begin{center}
\begin{tabular}{|l|c|c|c|c|c|c|c|}\hline
\multirow{2}{*}{Dataset} & \multirow{2}{*}{LDA} & Replicated & \multirow{2}{*}{fDARN} & \multirow{2}{*}{\docnade}& DeepDN& DeepDN& DeepDN \\
        &        & Softmax   &      &       &(1layer) &(2layer)  & (3layer) \\
\hline \hline
20~News & 1091 & 953 & 917 &  896 & {\bf 835} &877&923\\ \hline
RCV1-v2  & 1437 & 988 & 724 &  742 & 579 &552&{\bf 539}\\\hline
\end{tabular}
\end{center}
\caption{Test perplexity per word for models with 50 topics. 
  The
  results for LDA and Replicated Softmax were taken from \citet{Salakhutdinov-NIPS2010-softmax}.
}
\label{tab:results}
\vspace{-.5cm}
\end{table}
Table~\ref{tab:results} gathers the perplexity per word results for 20~Newsgroups and RCV1-V2. Theere we compare 5 different models: the Latent Dirichlet Allocation (LDA) \citep{BleiD2003}, the Replicated Softmax, the recent fast Deep AutoRegressive Networks (fDARN) \citep{mnih2014neural}, \docnade and Deep\docnade (DeepDN in the table). Each  model uses 50 latent topics. 
For the experiments with Deep\docnade, we provide the performance when using 1, 2, and 3 hidden layers. As shown in Table~\ref{tab:results}, Deep\docnade provides the best generative performances. Our best Deep\docnade models were trained with the Adam optimizer~\citep{kingma2014adam} and with the tanh activation function. The hyper-parameters of Adam were selected on the validation set. 

Crucially, following \cite{Uria2013b}, an ensemble approach is used to compute the probability of documents, where each component of the ensembles are the same Deep\docnade model evaluated on a different word ordering. Specifically, the perplexity with $M$ ensembles becomes as follows:
\begin{align}
    \exp\left(- \frac{1}{T}\sum_t \frac{1}{|\vv^t|}\log \left(\frac{1}{M}\sum_m p(\vv^{(t,m)})\right)\right),
\end{align}
where $T$ is the total number of examples, $M$ is the number of ensembles (word orderings) and $\vv^{(t,m)}$ denotes the $m^{\rm th}$ word ordering for the $t^{\rm th}$ documnet. We try $M=\left \{ 1, 2, 4, 16, 32, 64, 128, 256\right \}$, with the results in Table~\ref{tab:results} using $M=256$. For the 20~Newsgroups dataset, adding more hidden layers to Doc\docnade fails to provide further improvements. We hypothesize that the relatively small size of this dataset makes it hard to successfully train a deep model. However, the opposite is observed on the RCV1-V2 dataset, which is more than an order of magnitude larger than 20~Newsgroups. In this case, Deep\docnade outperforms fDARN and \docnade, with a relative perplexity reduction of 20\%, 24\% and 26\% with respectively 1,2 and 3 hidden layers. 

\begin{figure}[t]
\begin{center}
\begin{minipage}{.48\textwidth}
\includegraphics[width=\textwidth]{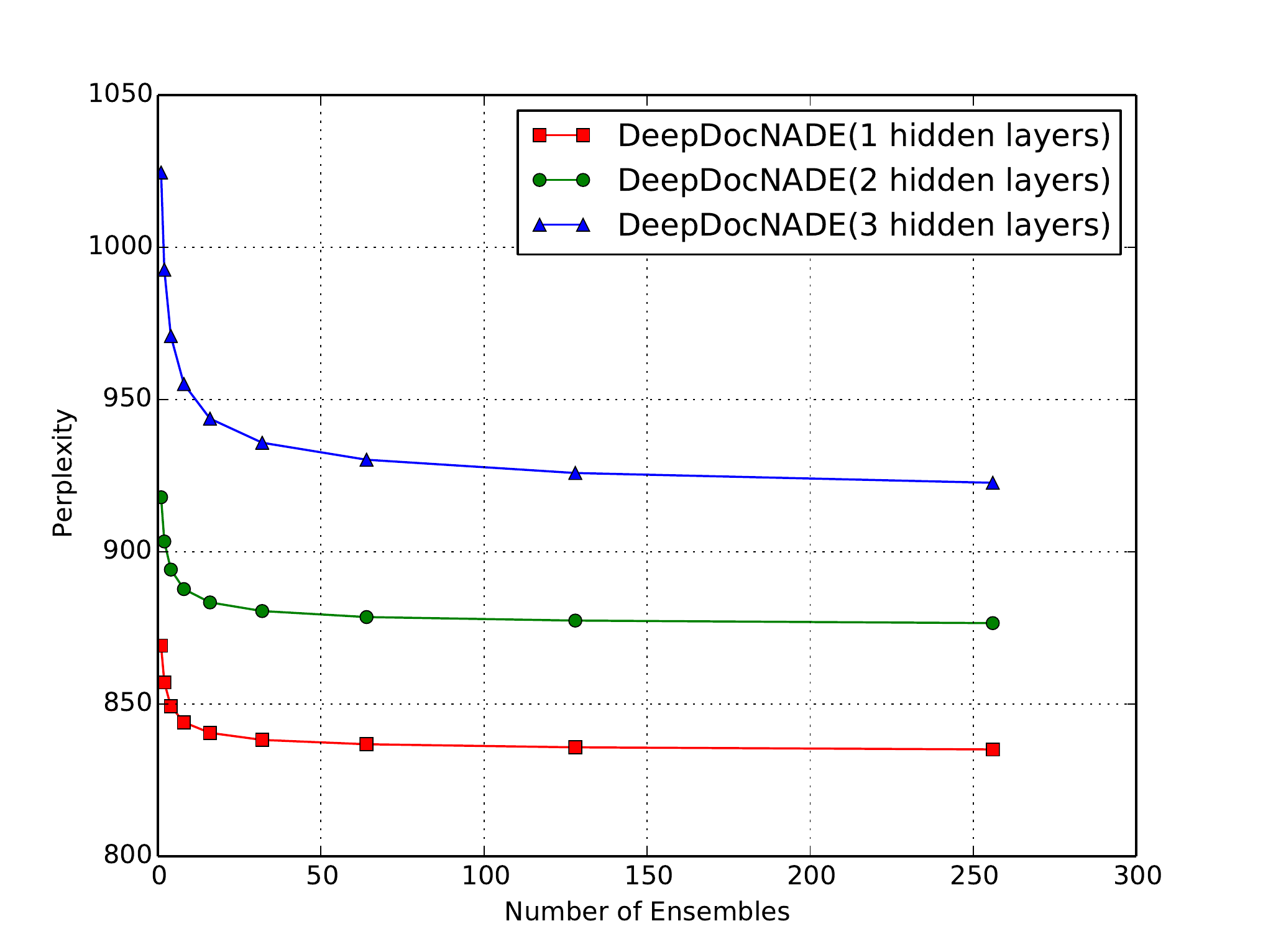}
\end{minipage}
\hspace*{0.2cm}
\begin{minipage}{.48\textwidth}
\includegraphics[width=\textwidth]{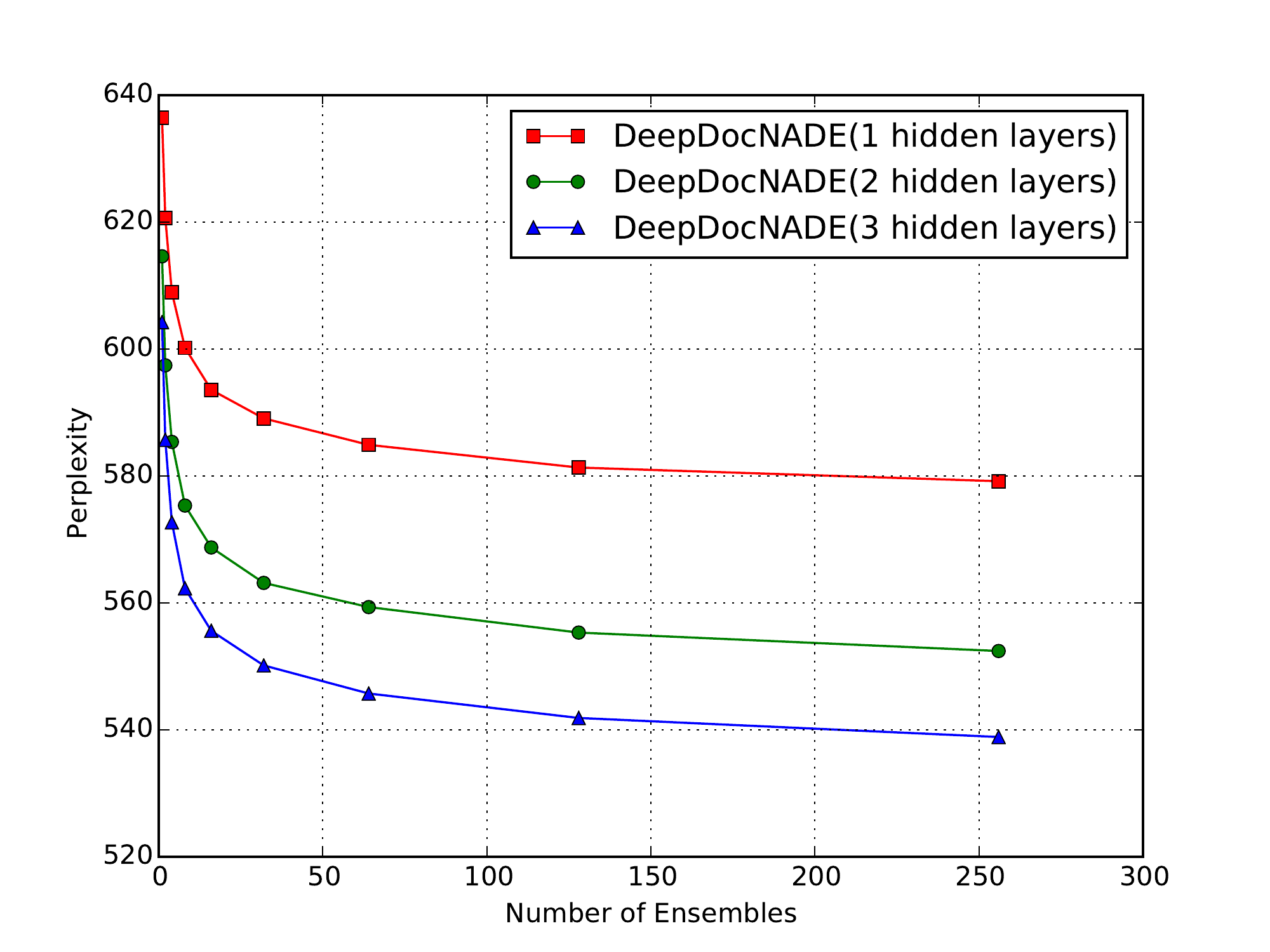}
\end{minipage}\\
\end{center}
\vspace*{-0.2cm}
\caption{Perplexity obtained with different numbers of word orderings $m$ for 20~Newsgroups on the left and RCV1-V2 on the right.}
\label{fig:ensemble}
\end{figure}

To illustrate the impact of $M$ on the performance of Deep\docnade, Figure~\ref{fig:ensemble} shows the perplexity on both datasets using the different values for $M$ that we tried. We can observe that beyond $M=128$, this hyper-parameter has only a minimal impact on the perplexity.

\vspace*{-0.2cm}

\subsection{Document Retrieval Evaluation}

A document retrieval evaluation task was also used to evaluate the quality of the document representation learned by each model. 
As in the previous section, the datasets under consideration are 20~Newsgroups and RCV1-V2. The experimental setup is the same for the 20~Newsgroups dataset, while for the RCV1-V2 dataset, we reproduce the same setup as the one used in~\citet{srivastava2013modeling}, where  the training set contained 794,414 examples and 10,000 examples constituted the test set. 

For \docnade and Deep\docnade, the representation of a document is obtained simply by computing the top-most hidden layer when feeding all words as input. 

The retrieval task follows the setup of \cite{srivastava2013modeling}. The documents in the training and validation sets are used as the database for retrieval, while the test set is used as the query set. The similarity between a query and all examples in the database is computed using the cosine similarity between their vector representations. For each query, documents in the database are then ranked according to this similarity, and precision/recall (PR) curves are computed, by comparing the label of the query documents with those of the database documents. Since documents sometimes have multiple labels (specifically those in RCV1-V2), for each query the PR curves for each of its labels are computed individually and then averaged. Finally, we report the global average of these (query-averaged) curves to compare models against each other. Training and model selection is otherwise performed as in the generative modeling evaluation.

\begin{figure}[h]
\begin{center}
20 NewsGroups\\
\begin{minipage}{.48\textwidth}
\includegraphics[width=\textwidth]{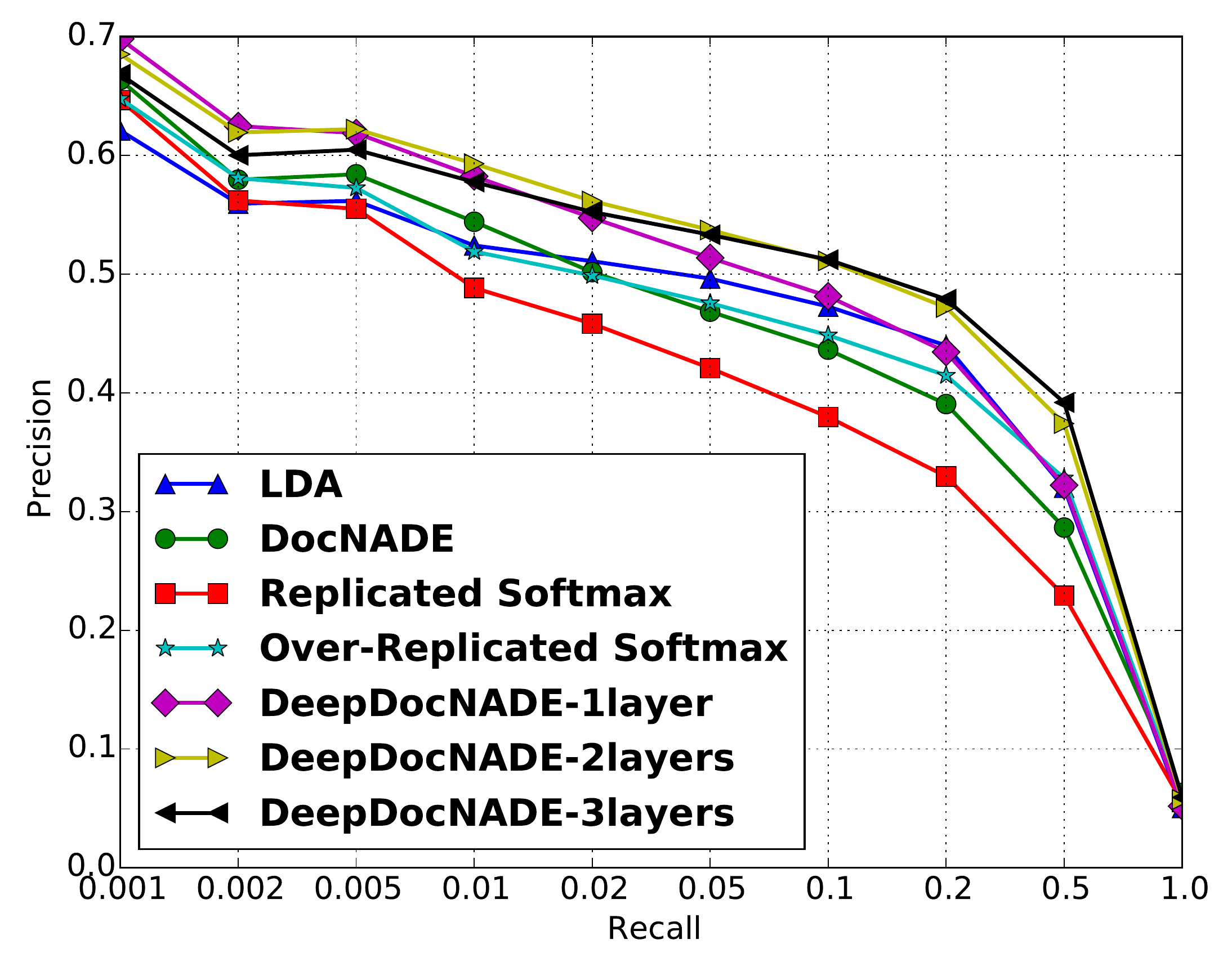}
\end{minipage}
\hspace*{0.2cm}
\begin{minipage}{.48\textwidth}
\includegraphics[width=\textwidth]{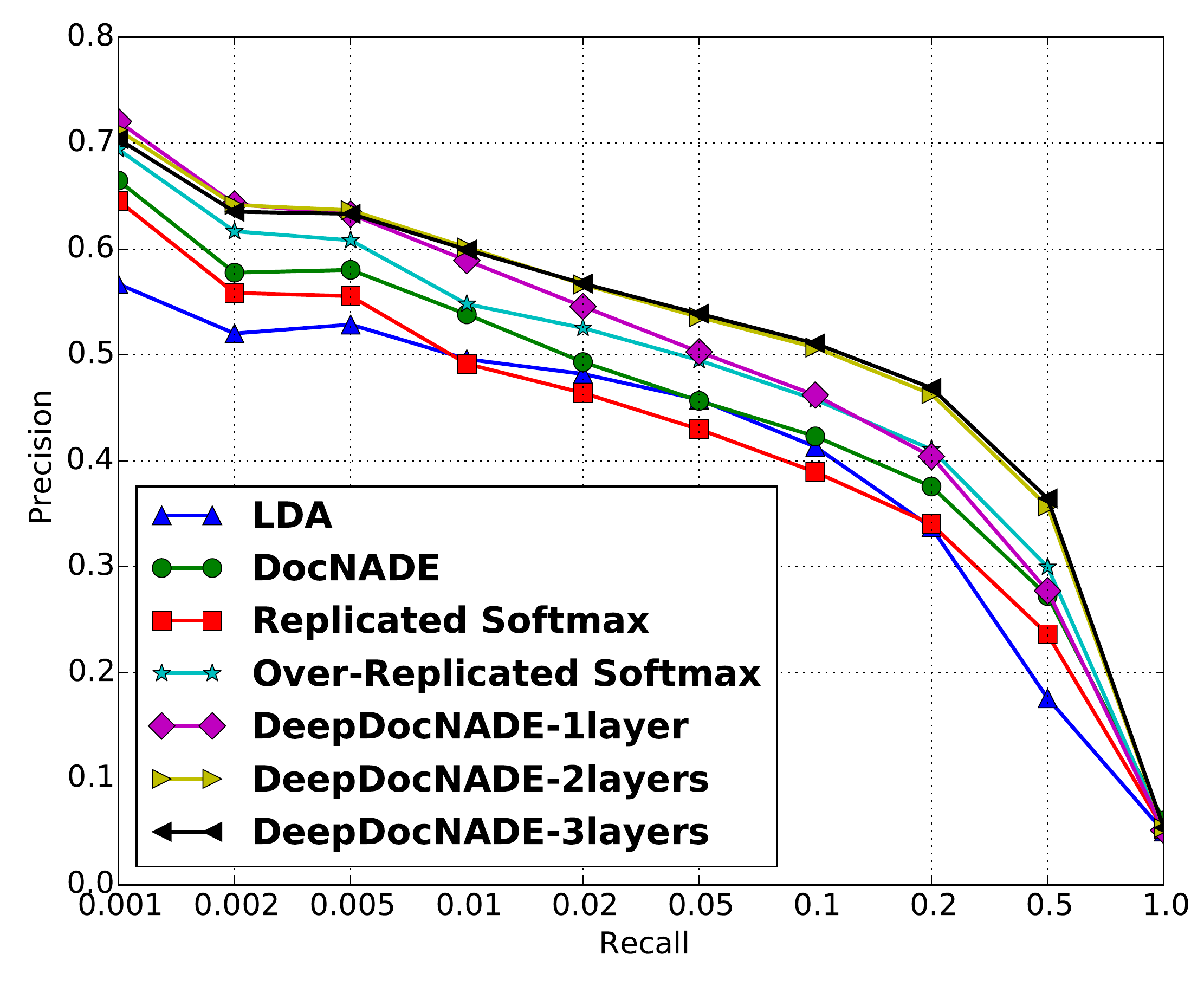}
\end{minipage}\\
Reuters RCV1-V2\\
\begin{minipage}{.48\textwidth}
\includegraphics[width=\textwidth]{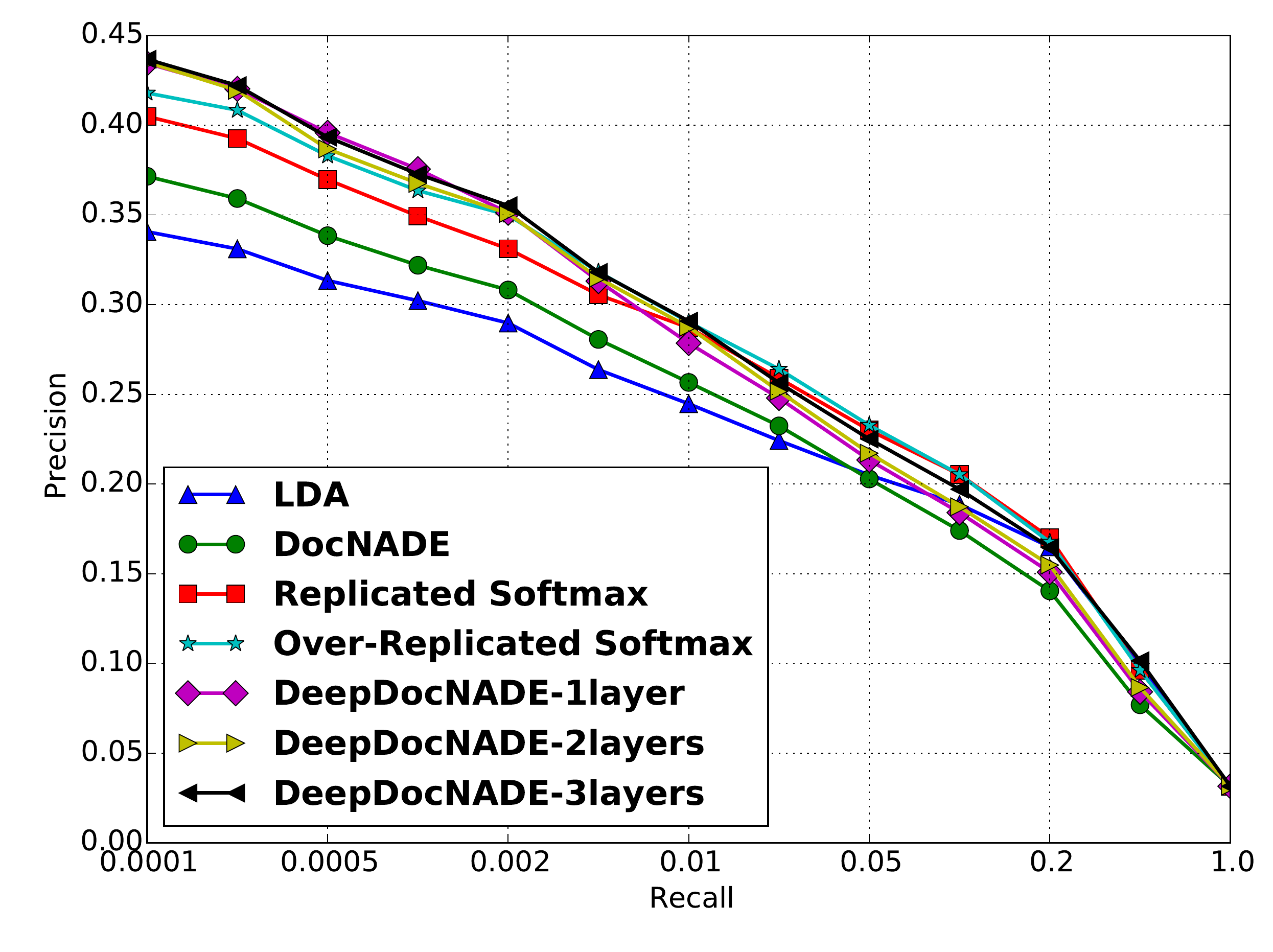}
\end{minipage}
\hspace*{0.2cm}
\begin{minipage}{.48\textwidth}
\includegraphics[width=\textwidth]{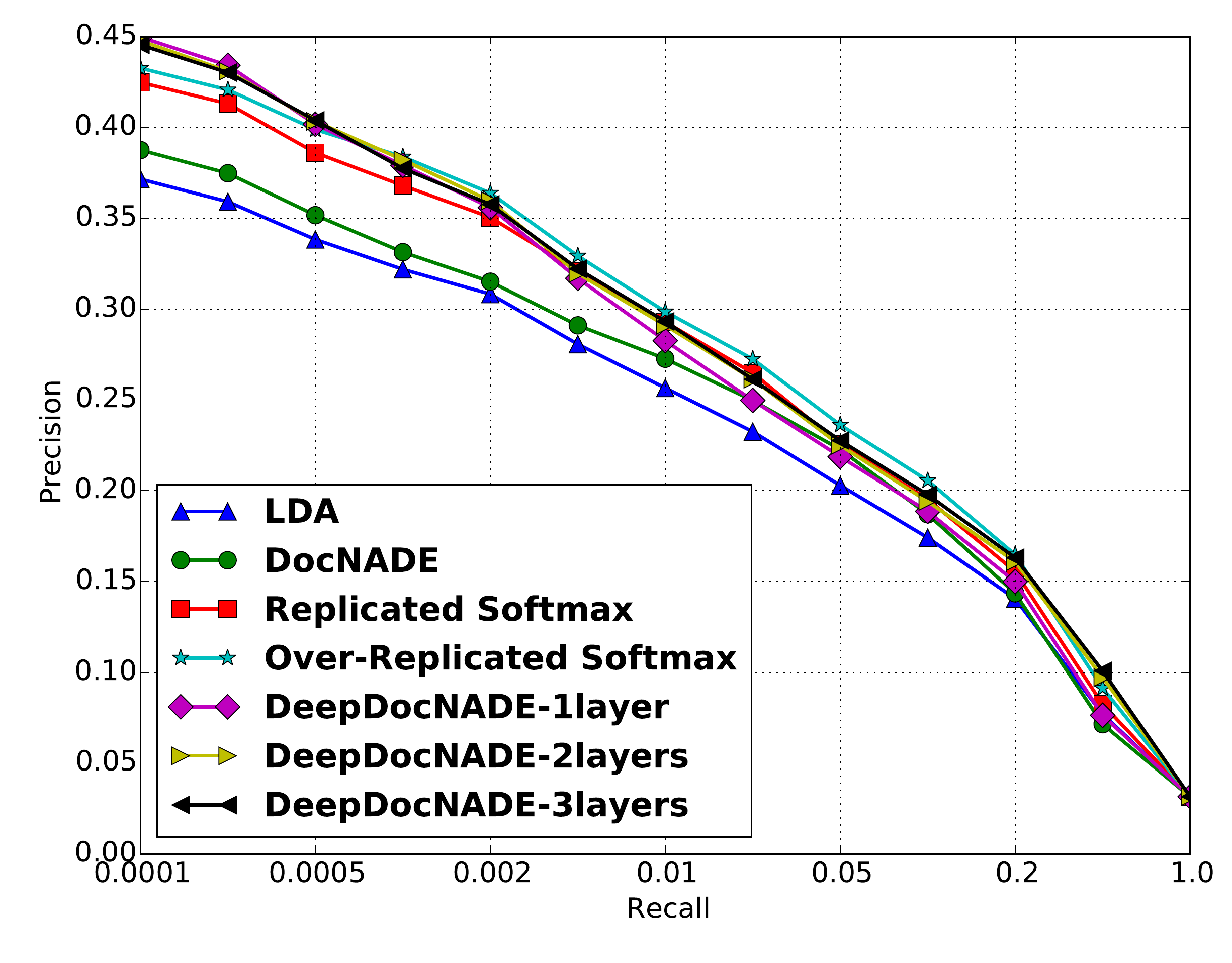}
\end{minipage}
\end{center}
\vspace*{-0.2cm}
\caption{Precision-Recall curves for document retrieval task. On the left are the results using a hidden layer size of 128, while the plots on the right are for a size of 512.}
\label{fig:x}
\end{figure}

As shown in Figure~\ref{fig:x}, Deep\docnade  always yields very competitive results,  on both datasets, and outperforming the other models in most cases. Specifically, for the 20~Newsgroups dataset, Deep\docnade with 2 and 3 hidden layers always perform better than the other methods. Deep\docnade with 1 hidden layer also performs better than the other baselines when retrieving the top few documents ( e.g.\ when recall is smaller than 0.2).


\vspace*{-0.2cm}

\subsection{Qualitative Inspection of Learned Representations}

In this section, we want to assess if the \docnade approach for topic modeling can capture meaningful semantic properties of texts. 

First, one way to explore the semantic properties of trained models is through their learned word embeddings. Each of the columns of the matrix $W$ represents a word $w$ where $W_{:,w}$ is the vector representation of $w$. Table~\ref{tab:results_topics} shows for some chosen words the five nearest words according to their embeddings, foro a \docnade model. We can observe for each example the semantic consistency of the word representations. Similar results can be observed for Deep\docnade models.


\begin{table*}[t]
\label{wordNeighbors}
\begin{center}
\begin{tabular}{|l|l|l|l|l|l|l|}
\hline
{\bf weapons } & {\bf medical } & {\bf companies } & {\bf define } & {\bf israel } & {\bf book } & {\bf windows }\\
\hline
weapon & treatment & demand & defined & israeli & reading & dos\\
shooting & medecine & commercial & definition & israelis & read & microsoft\\
firearms & patients & agency & refer & arab & books & version\\
assault & process & company & make & palestinian & relevent & ms\\
armed & studies & credit & examples & arabs & collection & pc\\
\hline
\end{tabular}
\end{center}
\caption{The five nearest neighbors in the word representation space learned by \docnade. 
}
\end{table*}

We've also attempted to measure whether the hidden units of the first hidden layer of \docnade and Deep\docnade models modeled distinct topics. Understanding the function represented by hidden units in neural networks is not a trivial affair, but we considered the following, simple approach. For a given hidden unit, its connections to words were interpreted as the importance of the word for the associated topic. Therefore, for a hidden unit, we selected the words having the strongest positive connections, i.e.\  for the $i_{th}$ hidden unit we chose the words $w$ that have the highest connection values $W_{i,w}$. 

With this approach, four topics were obtained from a \docnade model using the sigmoid activation function and trained on 20~Newsgroups, as shown in Table~\ref{tab:results_topics} and can be readily  interpreted as topics representing religion, space, sports and security. Note that those four topics are actually (sub)categories in 20 Newsgroups.

That said, we've had less success understanding the topics extracted when using the tanh activation function or when using Deep\docnade. It thus seems that these models are then choosing to learn a latent representation that isn't aligning its dimensions with concepts that are easily interpretable, even though it is clearly capturing well the statistics of documents (since our qualitative results with Deep\docnade are excellent).

\begin{table}[t]
\begin{center}
\begin{tabular}{|c|c|c|c|}\hline
\multicolumn{4}{|c|}{\bf Hidden unit topics}\\\hline
jesus         & shuttle    & season  & encryption \\     
atheism       & orbit      & players & escrow \\  
christianity  & lunar      & nhl     & pgp  \\
christ        & spacecraft & league  & crypto\\
athos         & nasa       & braves  & nsa\\
atheists      & space      & playoffs& rutgers\\
bible         & launch     & rangers & clipper\\
christians    & saturn     & hockey  & secure \\
sin           & billion    & pitching& encrypted\\
atheist       & satellite  & team    & keys\\\hline
\end{tabular}
\end{center}
\caption{Illustration of some topics learned by \docnade. A topic $i$ is visualized by picking the 10 words $w$ with strongest connection $W_{i w}$.}
\label{tab:results_topics}
\vspace{-.5cm}
\end{table}

\section{Language Modeling Experiments}

In this section, we test whether our proposed approach to incorporating a \docnade component to a neural language model can improve the performance of a neural language model. Specifically, we considered treating a text corpus as a sequence of documents. We used the APNews dataset, as provided by  \citet{Mnih+Hinton-2009}. Unfortunately, information about the original segmentation into documents of the corpus wasn't available in the data as provided by \citet{Mnih+Hinton-2009}, thus we simulated the presence of documents by grouping one or more adjacent sentences, for training and evaluating \docnade-LM, making sure the generated documents were non-overlapping. Thankfully, this approach still allows us to test whether \docnade-LM is able to effectively leverage the larger context of words in making its predictions. 

Since language models are generative models, the perplexity measured on some held-out texts provides an intrinsic and widely used evaluation criterion. Following \citet{mnih2007three} and \citet{Mnih+Hinton-2009},  we used the APNews dataset containing Associated Press news stories from 1995 and 1996. The dataset is again split into training, validation and test sets, with respectively 633,143, 43,702 and 44,601 sentences. The vocabulary is composed of 17,964 words.  A 100 dimensional feature vectors are used for these experiments. The validation set is used for model selection and the perplexity scores of Table~\ref{tab:results2} are computed on the test set.

\begin{table}[t]
\begin{center}
\begin{tabular}{|l|c|c|}\hline
Models & Number of grouped sentences & Perplexity\\
\hline \hline
KN5 & - & 123.2\\\hline
KN6 & - & 123.5\\\hline
LBL & - & 117.0\\\hline
HLBL & - & 112.1\\\hline
FFN & - & 119.78\\\hline
\docnade-LM  & 1 & 111.93\\\hline
\docnade-LM  & 2 & 110.9\\\hline
\docnade-LM  & 3 & 109.8\\\hline
\docnade-LM  & 4 & 109.78\\\hline
\docnade-LM  & 5 & 109.22\\\hline
\end{tabular}
\end{center}
\caption{Test perplexity per word for models with 100 topics. The results for HLBL and LBL were taken from \citet{Mnih+Hinton-2009}.
}
\label{tab:results2}
\vspace{-.5cm}
\end{table}

The FFN model in Table~\ref{tab:results2} corresponds to a regular neural (feed-forward) network language model. It is equivalent to setting ${\bf h}^{\rm DN}_i(\vv_{<i})$ to zero in Equation~\ref{eq:docnadelm-hidden}. These results are meant to measure whether the \docnade part of \docnade-LM can indeed help to improve performances.

We also compare to the log-bilinear language (LBL) model of \cite{mnih2007three}). While for the FFN model we used a hierarchical softmax to compute the conditional word probabilities (see Section~\ref{sec:docnade-lm}), the LBL model uses a full softmax output layer that uses the same word representation matrix at the input and output. This latter model is therefore slower to train. Later, \citet{Mnih+Hinton-2009} also proposed adaptive approaches to learning a structured softmax layer, thus we also compare with their best approach. All aforementioned baselines are 6-gram models, taking in consideration the last 5 previous words to predict the next one. We also compare with more traditional 5-gram and 6-gram models using Kneser-Ney smoothing, taken from \citet{mnih2007three}.

From Table~\ref{tab:results2}, we see that adding context to \docnade-LM, by increasing the size of the multi-sentence segments, significantly improves the performance of the model (compared to FFN) and also surpasses the performance of the most competitive alternative, the HLBL model.

\subsection{Qualitative Inspection of Learned Representations}

In this section we explore the semantic properties of texts learned by the \docnade-LM model. Interestingly, we can examine the two different components (DN and LM) separately. Because  the \docnade part and the language modeling part of the model each have their own word matrix, $\WW^{\rm DN}$ and  $\WW^{\rm LM}$ respectively, we can compare their contribution through these learned embeddings. As explained in the previous section, each of the columns of the matrices represents a word $w$ where $\WW^{\rm DN}_{:,w}$ and $\WW^{\rm LM}_{:,w}$ are two different vector representations of the same word $w$.

We can see by observing Tables~\ref{wordNeighbors_DN} and~\ref{wordNeighbors_time} that the two parts of the \docnade-LM model have learned different semantic properties of words. An interesting example is seen in the nearest neighbors of the word {\it israel}, where the \docnade focuses on the politico-cultural relation between these words, whereas the language model part seems to have learned the concept of countries in general.

\begin{table}[t]
\begin{center}
\begin{tabular}{|l|l|l|l|l|l|l|}
\hline
{\bf weapons } & {\bf medical } & {\bf companies } & {\bf define } & {\bf israel } & {\bf book } & {\bf windows }\\
\hline
security & doctor & industry & spoken & israeli & story & yards\\
humanitarian & health & corp & to & palestinian & novel & rubber\\
terrorists & medicine & firms & think\_of & jerusalem & author & piled\\
deployment & physicians & products & of & lebanon & joke & fire\_department\\
diplomats & treatment & company & bottom\_line & palestinians & writers & shell\\
\hline
\end{tabular}
\end{center}
\caption{The five nearest neighbors in the word representation space learned by the \docnade part of the \docnade-LM model. 
}
\label{wordNeighbors_DN}

\end{table}
\begin{table}[t]
\begin{center}
\begin{tabular}{|l|l|l|l|l|l|l|}
\hline
{\bf weapons } & {\bf medical } & {\bf companies } & {\bf define } & {\bf israel } & {\bf book } & {\bf windows }\\
\hline
systems & special & countries & talk\_about & china & film & houses\\
aircraft & japanese & nations & place & russia & service & room\\
drugs & bank & states & destroy & cuba & program & vehicle\\
equipment & media & americans & show & north\_korea & movie & restaurant\\
services & political & parties & over & lebanon & information & car\\
\hline
\end{tabular}
\end{center}
\caption{The five nearest neighbors in the word representation space learned by the language model part of the \docnade-LM model. 
}
\label{wordNeighbors_time}

\end{table}

\section{Conclusion}

We have presented models inspired by NADE that can achieve state-of-the-art performances for modeling documents.

Indeed, for topic modeling, \docnade had competitive results while its deep version, Deep\docnade, outperformed the current state-of-the-art in generative document modeling, based on test set perplexity. The similarly good performances where observed when we used these models as feature extractors to represent documents for the task of information retrieval. 

As for language modeling, the competitive performances of the DocNADE language models showed that combining contextual information by leveraging the \docnade neural network architecture can significantly improve the performance of a neural probabilistic N-gram language model.



\vskip 0.2in
\bibliography{DocNADE_bib}

\end{document}